\documentclass[journal]{IEEEtran}

\ifCLASSINFOpdf
\else
\fi

\usepackage{times}
\usepackage{epsfig}
\usepackage{graphicx}
\usepackage{amsmath}
\usepackage{amssymb}
\usepackage{multirow}

\usepackage{cite}
\usepackage{enumerate}
\usepackage{amsthm}
\usepackage{algorithm}
\usepackage{algorithmic}
\usepackage{cases}
\usepackage{bm}

\usepackage{amsthm}
\usepackage{cases}
\usepackage{bm}
\usepackage{algorithm}
\usepackage{algorithmic}
\usepackage{cases}
\usepackage{multirow}
\usepackage{color} 
\usepackage{enumerate}
\usepackage{etoolbox}
\usepackage{boxedminipage} 
\usepackage{booktabs} 
\usepackage{empheq} 
\usepackage{tikz} 
\usepackage{cases}
\usepackage{bbm}
\usepackage{tabularx}
\usepackage{enumitem}

\DeclareFontFamily{OT1}{pzc}{}
\DeclareFontShape{OT1}{pzc}{m}{it}{<-> s * [1.10] pzcmi7t}{}
\DeclareMathAlphabet{\mathpzc}{OT1}{pzc}{m}{it}

\newcommand{\MyMapTemplatePrefix}[4]{\expandafter#1\csname#3#4\endcsname{#2{#4}}}
\newcommand{\MyMapTemplatePrefixNew}[5]{\expandafter#1\csname#4#5\endcsname{#2{#3{#5}}}}
\forcsvlist{\MyMapTemplatePrefix {\def} {\mathbf} {}} {A,B,C,D,E,F,G,H,I,J,K,L,M,N,O,P,Q,R,S,T,U,V,W,X,Y,Z}
\forcsvlist{\MyMapTemplatePrefix {\def} {\mathbf} {}} {a,b,c,d,e,f,g,h,i,j,k,l,m,n,o,p,q,r,s,t,u,v,w,x,y,z,1,0}
\forcsvlist{\MyMapTemplatePrefix {\def} {\widetilde} {wt}} {A,B,C,D,E,F,G,H,I,J,K,L,M,N,O,P,Q,R,S,T,U,V,W,X,Y,Z}
\forcsvlist{\MyMapTemplatePrefix {\def} {\widetilde} {wt}} {a,b,c,d,e,f,g,h,i,j,k,l,m,n,o,p,q,r,s,t,u,v,w,x,y,z} 
\forcsvlist{\MyMapTemplatePrefixNew {\def} {\widetilde}{\mathbf} {tb}} {A,B,C,D,E,F,G,H,I,J,K,L,M,N,O,P,Q,R,S,T,U,V,W,X,Y,Z}
\forcsvlist{\MyMapTemplatePrefixNew {\def} {\widetilde}{\mathbf} {tb}} {a,b,c,d,e,f,g,h,i,j,k,l,m,n,o,p,q,r,s,t,u,v,w,x,y,z}
\forcsvlist{\MyMapTemplatePrefixNew {\def} {\overleftarrow}{\mathbf}{ol}} {A,B,C,D,E,F,G,H,I,J,K,L,M,N,O,P,Q,R,S,T,U,V,W,X,Y,Z}
\forcsvlist{\MyMapTemplatePrefixNew {\def}{\overrightarrow}{\mathbf}{or}} {A,B,C,D,E,F,G,H,I,J,K,L,M,N,O,P,Q,R,S,T,U,V,W,X,Y,Z}
\forcsvlist{\MyMapTemplatePrefix {\def} {\widehat} {wh}} {A,B,C,D,E,F,G,H,I,J,K,L,M,N,O,P,Q,R,S,T,U,V,W,X,Y,Z}
\forcsvlist{\MyMapTemplatePrefix {\def} {\widehat} {wh}} {a,b,c,d,e,f,g,h,i,j,k,l,m,n,o,p,q,r,s,t,u,v,w,x,y,z}
\forcsvlist{\MyMapTemplatePrefixNew {\def} {\widehat}{\mathbf} {hb}} {A,B,C,D,E,F,G,H,I,J,K,L,M,N,O,P,Q,R,S,T,U,V,W,X,Y,Z}
\forcsvlist{\MyMapTemplatePrefixNew {\def} {\widehat}{\mathbf} {hb}} {a,b,c,d,e,f,g,h,i,j,k,l,m,n,o,p,q,r,s,t,u,v,w,x,y,z}
\forcsvlist{\MyMapTemplatePrefix {\def} {\mathcal}{mc}} {A,B,C,D,E,F,G,H,I,J,K,L,M,N,O,P,Q,R,S,T,U,V,W,X,Y,Z}
\forcsvlist{\MyMapTemplatePrefix {\def} {\mathpzc}{mc}} {a,b,c,d,e,f,g,h,i,j,k,l,m,n,o,p,q,r,s,t,u,v,w,x,y,z}
\forcsvlist{\MyMapTemplatePrefixNew {\def} {\widetilde}{\mathcal} {tc}} {A,B,C,D,E,F,G,H,I,J,K,L,M,N,O,P,Q,R,S,T,U,V,W,X,Y,Z}
\forcsvlist{\MyMapTemplatePrefix {\def} {\mathbb} {mb}} {A,B,C,D,E,F,G,H,I,J,K,L,M,N,O,P,Q,R,S,T,U,V,W,X,Y,Z}
\forcsvlist{\MyMapTemplatePrefix {\DeclareMathOperator} {} {} } {tr,diag,sgn}

\def\tp{^\intercal}

\def\id{\text{id}} \def\ra{\text{ra}} \def\bep{\bm{\epsilon}} 
\def\bid{{\bar{\text{id}}}}
\def\tquery{{\text{qry}}}
\def\tkey{{\text{key}}}
\def\tvalue{{\text{val}}}

\definecolor{darkpastelgreen}{rgb}{0.01, 0.75, 0.24}
\definecolor{amethyst}{rgb}{0.6, 0.4, 0.8}
\definecolor{applegreen}{rgb}{0.55, 0.71, 0.0}
\definecolor{bittersweet}{rgb}{1.0, 0.44, 0.37}
\definecolor{blueviolet}{rgb}{0.54, 0.17, 0.89}
\definecolor{brightgreen}{rgb}{0.4, 1.0, 0.0}
\definecolor{brightpink}{rgb}{1.0, 0.0, 0.5}
\definecolor{caribbeangreen}{rgb}{0.0, 0.8, 0.6}
\definecolor{electricgreen}{rgb}{0.0, 1.0, 0.0}
\definecolor{rybgreen}{rgb}{0.4, 0.69, 0.2}

\graphicspath{figs/}

\usepackage{xcolor}

\usepackage{cite}
\usepackage[caption=false]{subfig}
\begin{document}
\bstctlcite{IEEEexample:BSTcontrol}
\title{Learning Fair Face Representation With Progressive Cross Transformer}
%
%
%
%

\author{Yong Li,
	Yufei Sun,
	Zhen Cui,
	Shiguang Shan,
	Jian Yang
	\thanks{Corresponding author: Zhen Cui, zhen.cui@njust.edu.cn}
	\thanks{Yong Li, Yufei Song, Zhen Cui, Jian Yang are with the Key Laboratory of Intelligent Perception and Systems for High-Dimensional Information, Ministry of Education, School of Computer Science and Engineering, Nanjing University of Science and Technology, Nanjing, China 210094 (e-mail: (yong.li, laolingjie, zhen.cui)@njust.edu.cn, csjyang@mail.njust.edu.cn).}
	\thanks{Shiguang Shan is with the Key Laboratory of Intelligent Information Processing of Chinese Academy of Sciences, Institute of Computing Technology, CAS, Beijing 100190, China, and with the University of Chinese Academy of Sciences, Beijing 100049, China, and also with CAS Center for Excellence in Brain Science and Intelligence Technology (e-mail: sgshan@ict.ac.cn).}
}

\markboth{Journal of \LaTeX\ Class Files,~Vol.~14, No.~8, August~2018}%
{Shell \MakeLowercase{\textit{et al.}}: Bare Demo of IEEEtran.cls for IEEE Journals}


\IEEEtitleabstractindextext{%
\begin{abstract}
Face recognition (FR) has made extraordinary progress owing to the advancement of deep convolutional neural networks. However, demographic bias among different racial cohorts still challenges the practical face recognition system.
The race factor has been proven to be a dilemma for fair FR (FFR) as the subject-related specific attributes induce the classification bias whilst carrying some useful cues for FR. To mitigate racial bias and meantime preserve robust FR, we abstract face identity-related representation as a signal denoising problem and propose a progressive cross transformer (PCT) method for fair face recognition. Originating from the signal decomposition theory, we attempt to decouple face representation into i) identity-related components and ii) noisy/identity-unrelated components induced by race. As an extension of signal subspace decomposition, we formulate face decoupling as a generalized functional expression model to cross-predict face identity and race information. The face expression model is further concretized by designing dual cross-transformers to distill identity-related components and suppress racial noises. In order to refine face representation, we take a progressive face decoupling way to learn identity/race-specific transformations, so that identity-unrelated components induced by race could be better disentangled. We evaluate the proposed PCT on the public fair face recognition benchmarks (BFW, RFW) and verify that PCT is capable of mitigating bias in face recognition while achieving state-of-the-art FR performance. Besides, visualization results also show that the attention maps in PCT can well reveal the race-related/biased facial regions.
\end{abstract}

}

\maketitle

\IEEEdisplaynontitleabstractindextext

\IEEEpeerreviewmaketitle

\ifCLASSOPTIONcompsoc
\IEEEraisesectionheading{\section{Introduction}\label{sec:introduction}}
\else
\section{Introduction}
\label{sec:introduction}
\fi

\IEEEPARstart{A}{utomatic} face recognition has achieved considerable success with the rapid developments of deep learning algorithms \cite{schroff2015facenet, liu2017sphereface, wang2018deep, wang2018cosface, huang2020improving}. However, Face recognition (FR) systems are found to exhibit discriminatory behaviors against certain demographic groups \cite{klare2012face, drozdowski2020demographic, grother2019face, wang2020mitigating, lu2019experimental, acien2018measuring}.
Every face not only reflects individual identity, but also exhibits many demographic attributes, such as race, ethnicity, age, gender and other visible forms of self-expression \cite{merler2019diversity}. The face recognition systems are expected to work equally accurate for each of us. It means the performance of a desired FR system should not vary for different individuals or demographic groups.
However, as reported in the 2019 NIST Face Recognition Vendor Test \cite{grother2019face}, all participating FR algorithms exhibit different levels of biased performances across various demographic (e.g., race, country of birth, gender, age) groups.
Such FR systems with bias against specific minorities can lead to unjust or prejudicial outcomes.
 To prevent the unexpected side effects and to ensure the long-term acceptance of the FR algorithms, the development and deployment of unbiased FR systems are vital and essential.

Currently, the popular FR models are with convolutional neural networks (CNNs) trained on large scale training data to represent face features in N-dimensions with minimal distances between the same identity and maximum between unique identities. During the training process, the bias is inevitably introduced by both the imbalanced training data and the mapping function that encodes the input face image into a low-dimensional vector.
Since the commonly used FR datasets (e.g., CASIA-WebFace  \cite{yi2014learning}, MS-Celeb-1M  \cite{guo2016ms}) are collected from the Internet, face images are naturally imbalanced in different demographic groups.
To address the data imbalance issue, data re-sampling methods \cite{zhang2009cost, huang2019deep, mullick2019generative} have been carefully exploited to manually balance the data distribution by under-sampling the majority or oversampling the minority classes. However, naively training on a balanced dataset is beneficial for fair face recognition but can still lead to bias to some degree \cite{zhang2020class}.

Considering the limited benefits brought by the balanced datasets, a number of FR approaches have been proposed \cite{wang2020mitigating, wang2019racial, gong2020jointly, robinson2020face} to mitigate the performance bias among different demographic groups. These methods aim to learn a deep neural network that can fairly encodes the input faces among different demographic groups, regardless of whether the training datasets are balanced or not. For example, Wang et al. \cite{wang2019racial} proposed a deep information maximization adaptation network by considering Caucasian as the source domain and other races as target domains.
Wang et al. \cite{wang2020mitigating} further proposed a reinforcement learning-based network to find the optimal margins for different racial groups.

Intuitively, in order to learn the fair face representation, the face recognition model needs to mitigate the bias caused by or related to race. This type of analysis comes up in multiple literature \cite{wang2019decorrelated, gong2020jointly}. Among them, Gong et al \cite{gong2020jointly} proposed a novel de-biasing adversarial network (DebFace) that learns to encode disentangled feature representations for both unbiased face recognition and attribute estimation. The learned disentangled face representations exhibit less bias among different demographic groups while show degraded FR performance simultaneously. This is because some demographic attributes such as race and gender are critical components that constitute face patterns and are useful cues for FR\cite{gong2020jointly, liu2018exploring, lu2019experimental, 2020_AAAI_Discrimination_Serna}. With similar motivation, Gong et al \cite{gong2020mitigating} proposed a group adaptive classifier that mitigates FR bias with adaptive convolution kernels and attention mechanisms on faces according to the demographic attributes.

Different with previous fair FR methods that either transfer knowledge among different racial groups or learn the disentangled face representation, we aim to resolve this challenge from a signal denoising perspective.
Given an input face image, we aim to distill the identity-related face representation by suppressing the noisy/identity-unrelated representation induced by race. To this end, we attempt to decompose the input face image into two parts: 1) identity-related representation, and 2) noisy/identity-unrelated representation induced by race.
We abstract the process of face representation as a signal denoising problem based on the signal subspace theory, and derive a generalized functional expression model for face decoupling. To learn the face expression model, we further concretize the unsolved function with the transformer. Specifically, the dual cross-transformers are designed to bridge two streams of the encoding process, one is an identity-related distillation and the other is the identity-unrelated suppression induced by race.
In order to better refine face representation, we take a progressive face decoupling way to learn identity/race-specific transformation, so that identity-unrelated representation introduced by race could be disentangled gradually.
We finally frame the proposed method into the progressive cross transformer (PCT) framework for fair face recognition.
Concretely, PCT consists of several cross transformer (CT) modules, each CT module learns the ethnicity-specific transformations to distil the bias caused by their racial groups. The CT modules can be integrated into the deep FR networks in a top-down manner so that the face representation is transformed and de-biased gradually.
By merely disentangling the identity-unrelated component induced by race, our proposed PCT is capable of mitigating the racial bias while enhancing the discriminative ability of the learned face representation.

Our contributions can be summarized as follows:
\begin{itemize}
	\item To suppress racial bias in face representation, we analyze face decoupling from a signal denoising perspective and abstract the distillation process of face representation as a generalized functional expression model, which falls into the transformer framework.
	\item
	We propose a progressive cross transformer method for fair face recognition, which extends the standard transformer as the cross dual streams of transformers. The method mitigates the bias components from the identity representations by cross transforming the features of identity and race progressively.
	\item Extensive experiments on two popular race aware training datasets demonstrate the feasibility of our proposed PCT, and meantime report the state-of-the-art recognition accuracy with lower bias on different racial groups.
\end{itemize}

\section{Related Work} \label{sec:related_work}

\subsection{Fair face recognition}

A number of studies have revealed that face recognition algorithms perform unequally on different demographic attributes \cite{klare2012face, drozdowski2020demographic, grother2019face, wang2020mitigating, lu2019experimental, acien2018measuring}.
The 2002 NIST Face Recognition Vendor Test (FRVT) \cite{phillips2003face} is believed to be the first study that showed that non-deep FR algorithms show unequal FR accuracies among different races. The recent NIST Face Recognition Vendor Test in 2019 \cite{grother2019face} reveals that all participating FR algorithms exhibit unequal performances on demographic attributes, e.g., gender, race, and age groups. Since then, some representative FR datasets have been released to study the FR bias.
Merler et al. \cite{merler2019diversity} released the Diversity in Faces (DiF) dataset that contains annotations of 1 million human facial images to advance the study of fairness in FR. The annotations in DiF dataset were generated using ten facial coding schemes that provide human-interpretable quantitative measures of intrinsic facial features.
Recently, Wang et al. \cite{wang2019racial} contributed a popular Racial Faces in-the-Wild (RFW) dataset, on which they validated the racial bias of various state-of-the-art FR methods, and proposed the unsupervised domain adaptation approach to alleviate the racial bias.  RFW dataset has been served as a testbed to fairly measure the bias of various FR algorithms.
Afterward, Wang et al. \cite{wang2020mitigating} released the popular BUPT-Balancedface dataset  where face images are balanced in various races, and BUPT-Globalface dataset that reveals the real distribution of the world’s population.
Similarly, Robinson et al. \cite{robinson2020face} proposed the BFW benchmark dataset for evaluating the bias of current FR recognition systems. They reduced the performance gaps among different subgroups and showed a notable boost in overall FR performance by learning subgroup-specific thresholds.
In this paper, we adopt the BUPT-Balancedface and BUPT-Globalface datasets for training and evaluate our proposed algorithm on RFW and BFW datasets.

Recently, a lot of studies aim to learn fair face representation independent of the balanced training datasets \cite{cavazos2020accuracy, robinson2020face, wang2020mitigating, gong2020jointly, gong2020mitigating}.
Inspired by the observations that the FR error rates on non-Caucasians are usually much higher than that of Caucasians, Wang et al. \cite{wang2020mitigating} proposed to manually select margin for Caucasians and learn the optimal margins of the non-Caucasians by Q-learning.
Gong et al. \cite{gong2020jointly} introduced a debiasing adversarial network with four specific classifiers, in which one for identity classification and the other three for demographic attributes estimation. The disentangled face representation in \cite{gong2020jointly} shows more equal FR accuracies among various demographic groups and degraded FR performance at the same time. This is because some demographic attributes such as race and gender are critical components that constitute face patterns and are beneficial for FR \cite{gong2020jointly, liu2018exploring}.
Gong et al. \cite{gong2020mitigating} proposed a group adaptive classifier (GAC) that mitigates the FR bias by using adaptive convolution kernels and attention mechanisms on faces based on their demographic attributes. The adaptive convolution kernels and attention mechanisms in GAC help activate different facial regions for face identification and lead to more discriminative features according to their demographics.

\begin{figure*}[htb]
	\centering
	\includegraphics[width=1.0\linewidth]{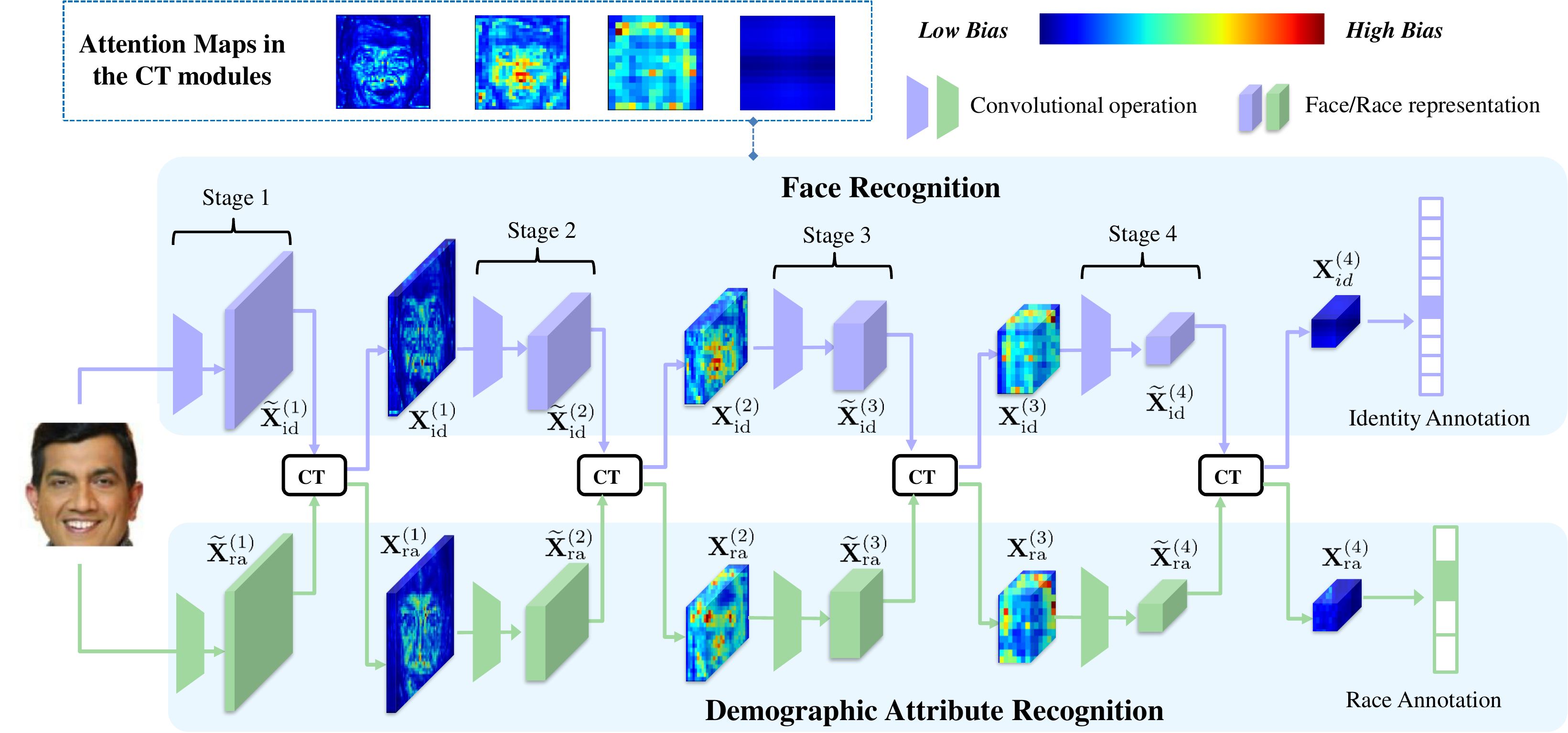}
	\caption{
		Main idea of the proposed progressive cross transformer (PCT). PCT takes a face image as input and obtains the stage-wise identity $\tbX^{(t)}_\id$ and race $\tbX^{(t)}_\ra$ representation at $t$-th stage. PCT then suppresses the racial bias from the input identity representation $\tbX^{(t)}_\id$ via the CT module and obtains the de-biased representation $\mathbf{X}^{(t)}_\id$. The learned attention maps in the stage-wise CT modules have been shown in the top rectangle, where the deep red means the biased and race-related facial regions. The attention maps mainly cover the facial texture, facial features, facial contour, which are race-related facial regions. More visual examples and explanations can be seen in Section \ref{sec:ablation_study}.
	}
	\label{fig:pct_framework}
\end{figure*}

\subsection{Transformer}

Transformer was proposed in \cite{waswani2017attention} for sequence to sequence tasks such as language translation, to replace traditional recurrent models. Transformer architecture has become the de-facto standard for natural language processing tasks.
 The main idea of the original transformer is to compute self-attention by comparing a feature to all other features in the sequence. In detail, features are first encoded to obtain query ($Q$) and memory (including key ($K$) and value ($V$)) embedding by linear projections. The product of the query $Q$ with keys $K$ is used as the attention weights for value $V$.
A position embedding is also exploited and added to these representations to incorporate the positional information which is lost in such a non-convolutional paradigm.
 Transformers are especially good at modeling long-range dependencies between elements of a sequence. Dosovitisky et al. \cite{dosovitskiy2020image} proposed the Vision Transformer (ViT) and showed the pure-transformer networks can achieve state-of-the-art results for image classification.
Since then, there have been many attempts to adapt transformers towards vision tasks including object detection \cite{carion2020end}, segmentation \cite{jin2021trseg, Wang_2021_CVPR}, image classification \cite{Srinivas_2021_CVPR}, multiple object tracking \cite{meinhardt2021trackformer, Chen_2021_CVPR}, video processing \cite{girdhar2019video, fang2020multi}.

For face-related tasks, Zhong et al. \cite{zhong2021face} investigated and verified the feasibility of applying transformer models for face recognition.
Jacob et al. \cite{Jacob_2021_CVPR}  proposed a transformer encoder architecture to capture the relationships between different facial action units for the wide range of facial expressions. The methods in \cite{li2021mvit, ma2021robust} translated the facial images into sequences of visual words and perform facial expression recognition from a global perspective.
Transformer has also been exploited to tackle the face clustering problem \cite {nguyen2021clusformer} thanks to its effective self-attention mechanism.

In this paper, we develop the progressive cross transformer (PCT) for fair face recognition.  The representations from the faces and demographics are cross transformed adaptively to learn and suppress the bias induced by race.
Unlike previous  transformer-based methods that learn a set of visual tokens and model the long-range dependencies via a self-attention mechanism, our proposed PCT implements  the cross-attention mechanism globally across the face and ethnicity representations in a top-down manner.
PCT enables the model to mitigate the ethnicity-biased information progressively and recognize the input faces equally.
We will formulate the transformation between the face and ethnicity representations for feature de-biasing from a signal subspace decomposition perspective in the next section.

\section{Motivation and Formulation}\label{sec:motivation}

An observed face signal $\tbx$ may consist of the pure identity signal $\x_\id$, noise signal $\bep_\bid$ and  measuring error $\e$. We assume that the identity signal $\x_\id$ comes from the system $\H$, i.e., $\x_\id=f_\id(\H,\bm\theta_H)$, where $\bm\theta_H$ is the estimated parameter, the noise signal $\bep_\bid$ is generated by the other non-id system $\S$, i.e., $\bep_\bid = f_\bid(\S,\bm\theta_S)$ with the system parameter $\bm\theta_S$, and the measuring error $\e$ is assumed to be additive white Gaussian noise. Formally, we can express the input face signal $\tbx$ as
\begin{align}
\tbx = \x_\id + \bep_\bid + \e = f_\id(\H,\bm\theta_H) + f_\bid(\S,\bm\theta_S) + \e.
\label{equ:id_noise}
\end{align}

In the case of the plain linear system, the basic vectors (w.r.t columns) of $\H$ ($\S$) are often assumed to be linearly independent, and accordingly the two corresponding spaces, Range($\H$) and Range($\S$), are disjoint and often non-orthogonal. Thus, we can rewrite Eqn.~(\ref{equ:id_noise}) as
\begin{align}
\tbx = \H\bm\theta_H + \S\bm\theta_S + \e.
\label{equ:id_noise_linear}
\end{align}
Suppose the additive white Gaussian noise $\e$ and the structure noise $\bep_\bid$ are orthogonal, then we have $\langle \S\bm\theta_S, \e\rangle = 0$, and further $\S\tp\e=0$ if $\bm\theta_S\neq 0$. Likewise, if the desired signal $\x_\id$ is orthogonal to $\e$, then we have $\H\tp\e=0$. We premultiply with the orthogonal projection matrix $\P_H^\bot$ to the space Range$(\H)$ in Eqn.~(\ref{equ:id_noise_linear}), and can derive as $\P_H^\bot\tbx = \P_H^\bot\S\bm\theta_S + \e$ by using $\P_H^\bot\H=0$ and $\P_H^\bot\e=\e$. Again, after premultiplication with the matrix $\S\tp$, we can reach $\bm\theta_S = (\S\tp\P_H^\bot\S)^{-1}\S\tp\P_H^\bot\tbx$. Therefore, we can express the noise signal as $\bep_\bid =\S\bm\theta_S = \S(\S\tp\P_H^\bot\S)^{-1}\S\tp\P_H^\bot\tbx = E_{S|H}\tbx$, where $E_{S|H}$ is the oblique projector from Range($\H$) to Range($\S$). In practice, the systems $\H$ and $\S$ are often unknown and also difficult to be estimated exactly. For fair face recognition we focus in this work, the only input signal $\tbx$ and some corresponding annotations (e.g., id and race information) can be used for training. Thus, it is intractable to decompose the input signal $\tbx$ into the latent identity signal $\x_\id$ and the non-id noise $\bep_\bid$. Moreover, the calculation has a high computation burden due to the inverse operation of matrix. In fact, the high-dimension signals usually lie in the non-linear space.

Now we rethink the estimation process of the noise signal $\bep_\bid$, and abstract it as a more generalized model with learnable parameters. Formally, we define $\bep_\bid \doteq \mcC(H, S)\cdot\mcP_H(\tbx)$ ($=[\S(\S\tp\P_H^\bot\S)^{-1}\S\tp][\P_H^\bot\tbx]$ in the linear system case), where $\mcP_H$ denotes the projection function of $\tbx$ to the space $H$ and $\mcC$ denotes the transformation function from the space $H$ to the space $S$. Due to the unknown spaces $H$ and $S$, we take the parameterizing strategy, and define the estimation process of noise signal $\bep_\bid$ as
\begin{align}
\bep_\bid \doteq \mcC_H(H(\tbx), S(\tbx))\cdot\mcP_S(\tbx),
\label{equ:epsilon_nonid}
\end{align}
where $H,S$ are estimated from the input signal $\tbx$, and $\mcC_S$ means the transformed target space is $S$.
Accordingly, we can define the identity signal $\x_\id$ as
\begin{align}
\x_\id \doteq  \mcC_S(H(\tbx), S(\tbx))\cdot\mcP_H(\tbx).
\label{equ:x_id}
\end{align}
So far, we have abstracted and modeled the basic decomposition process for the input signal $\tbx$, as formulated in Eqn.~(\ref{equ:epsilon_nonid}) and (\ref{equ:x_id}). However, the estimation processes of $\x_\id$ and $\bep_\bid$ should be mutually dependent based on Eqn.~(\ref{equ:id_noise}). If ignoring the measuring error $\e$, we can derive them as
\begin{align}
\x_\id \simeq \tbx - \bep_\bid,\quad
\bep_\bid \simeq \tbx-\x_\id,
\label{equ:id_and_noise}
\end{align}
which are more suit for the progressive stage learning through the interaction between $\x_\id$ and $\bep_\bid$ as used in Section~\ref{sec:PCT}.

Inspired by the success of transformer~\cite{waswani2017attention}, we may concretize the transformation functions $\mcC_S,\mcC_H$ and the projection functions $\mcP_S, \mcP_H$. Given the multi-channel feature $\tbX\in\mbR^{n\times d}$ with  spatial position $n=h\times w$ and feature dimension $d$, we define as follows,
\begin{align}
\mcC_S &\doteq \text{Softmax}\left(\frac{G_\id^\tquery(\tbX)\times [G_\bid^\tkey (\tbX)]\tp }{\sqrt{C}}\right),\label{equ:id_projection1}\\
\mcP_H & \doteq G_\id^\tvalue (\tbX),
\label{equ:id_projection2}
\end{align}
where $G_\id^\tquery$ means an identity transformation function w.r.t the query of transformer, $G_\bid^\tkey$ denotes a non-id function w.r.t the key of transformer, $G_\id^\tvalue$ is a identity projection function w.r.t the value of transformer, and $C$ is the normalization factor. Similarly, we can have
\begin{align}
\mcC_H &\doteq \text{Softmax}\left(\frac{G_\bid^\tquery(\tbX)\times [G_\id^\tkey (\tbX)]\tp }{\sqrt{C}}\right),\label{equ:race_projection1}\\
\mcP_S &\doteq G_\bid^\tvalue (\tbX).
\label{equ:race_projection2}
\end{align}
Hereby, we derive the learnable parameterized model with the transformer mechanism, where face identity signal can be decomposed from input face signal.

\section{Progressive Cross Transformer}\label{sec:PCT}

\subsection{The Network Framework}

Fig.~\ref{fig:pct_framework} shows the network framework of the proposed PCT.
To obtain the face and racial representation, the network structure of PCT is deployed to consist of two branches: 1) the top branch is used to learn the representation related to face identity, and 2) the bottom branch aims to capture identity-unrelated feature induced by race factor. For a convenient description, below we abuse the term ``race" as the identity-unrelated component because the identity-unrelated branch is constrained by the race factor. To extract robust feature, we take the typical face recognition network ResNet-34 \cite{he2016deep} as the infrastructure.  A ResNet typically consists of several stages and each stage consists of multiple bottleneck blocks with residual connections \cite{li2021graph}. By virtue of the multiple stages within the FR and RC networks, we are capable of extracting the top-down facial/racial representations based on the output of each stage within the FR and RC network, which facilitates progressive learning naturally.

Given an input face image, suppose we obtain the decoupled face representation $\tbX^{(t)}_\id \in \mbR^{n\times d}, \tbX^{(t)}_\bid \in \mbR^{n\times d}$ from the identity-related/unrelated branches at the $t$-th stage respectively. As the identity-unrelated feature is mainly induced by race factor, so we denote $\tbX^{(t)}_\bid$ with the race feature $\tbX^{(t)}_\ra$ for simplification. At the first stage, we may use the convolutional feature as the input to produce the preliminary face representations $\tbX^{(1)}_\id, \tbX^{(1)}_\ra$ with different convolution kernels.
The extracted features $\tbX^{(t)}_\id, \tbX^{(t)}_\ra$ are required for a further separation in the progressive distillation of face representation. To disentangle those identity-unrelated information, we design the cross transformer (CT) module to infer less biased facial representation by taking $\tbX^{(t)}_\id$, $\tbX^{(t)}_\ra$ as input, formally,
\begin{align}
\X^{(t)}_\id, \X^{(t)}_\ra &\leftarrow \text{CT}(\tbX^{(t)}_\id, \tbX^{(t)}_\ra),\\
\tbX^{(t+1)}_\id, \tbX^{(t+1)}_\ra &\leftarrow f(\X^{(t)}_\id), f(\X^{(t)}_\ra),\label{eqn:progres2}
\end{align}
where the cross transformer follows the basic idea as defined in Eqn.~(\ref{equ:epsilon_nonid})-(\ref{equ:race_projection2}) in Section~\ref{sec:motivation}, and $f$ is a function to match the input of the next stage. More details of the CT module are presented in Section~\ref{sec:CT}. With the evolution of face decoupling, more identity information $\tbX_\id$ will be aggregated into the identity-related branch while those identity-unrelated racial information is suppressed by the affect of the other branch of network.
At the last stage, we denote the output decoupled features with
\begin{align}
\X_\id, \X_\ra \leftarrow \X^{(T)}_\id, \X^{(T)}_\ra,
\end{align}
where $T$ is the number of progressive stages. For training, the decoupled representations $\X_\id, \X_\ra$ are fed into the constraint loss functions w.r.t face identity and face race labels. When testing, we only need perform the feeding-forward pass encoding to produce the representation related to face identity for fair face recognition.

\subsection{Cross Transformer Module}\label{sec:CT}

\begin{figure}[!htb]
	\centering
	\includegraphics[width=1.0\linewidth]{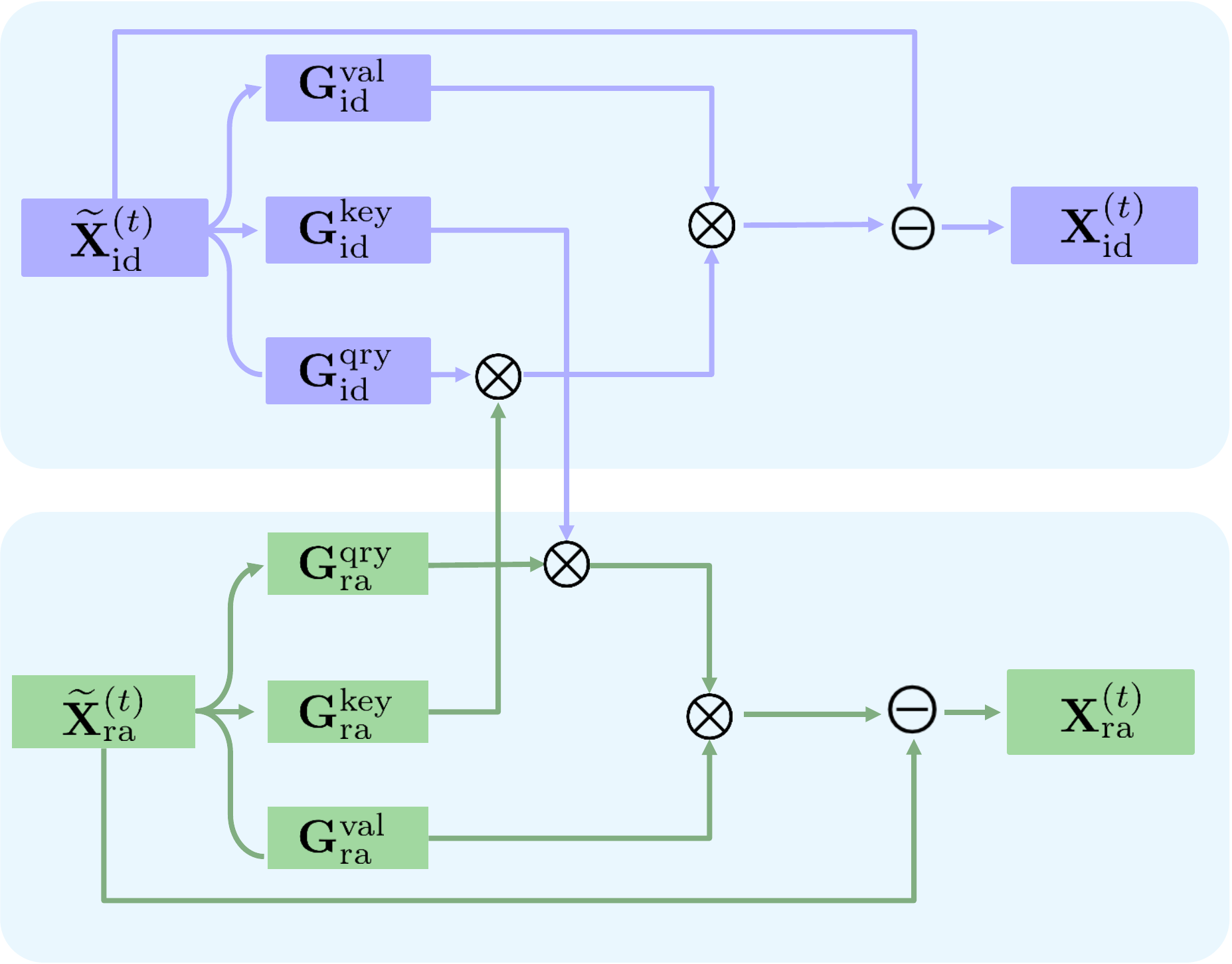}
	\caption{
	Illustration of proposed cross transformer (CT) module. The CT module takes $\tbX^{(t)}_\id$ and $\tbX^{(t)}_\ra$ as the input and output the decoupled resentations $\mathbf{X}^{(t)}_\id$ and $\mathbf{X}^{(t)}_\ra$. $\bigotimes$ means matrix multiplication. $\ominus$ means element-wise subtraction.
	}
	\label{fig:ct_unit_framework}
\end{figure}

Fig.~\ref{fig:ct_unit_framework} shows the main idea of the proposed CT module.
To suppress identity-unrelated component induced by race, this module at $t$-th stage takes $\tbX^{(t)}_\id$ and $\tbX^{(t)}_\ra$ as the input and outputs the purer representations $\mathbf{X}^{(t)}_\id$ and $\mathbf{X}^{(t)}_\ra$ as identity information and noises based on the transformer mechanism.

Formally, we concretize the definition in Eqns.~(\ref{equ:id_projection1})-(\ref{equ:race_projection2}). Following the basic idea of transformer, we define three learnable matrices related to identity transformation: $\W^\tkey_\id, \W^\tquery_\id, \W^\tvalue_\id \in \mathbb{R}^{d \times d'}$. Likewise, we also introduce other three learnable weights on the branch of race, i.e., $\W^\tkey_\ra, \W^\tquery_\ra, \W^\tvalue_\ra \in \mathbb{R}^{d \times d'}$. The function $G$ ($\G$ as value) in Eqns.~(\ref{equ:id_projection1})-(\ref{equ:race_projection2}) is rewritten as
\begin{align}
\G_\id^\tkey&= \tbX^{(t)}_\id \W_\id^\tkey+\b_\id^\tkey, &\G_\ra^\tkey= \tbX^{(t)}_\ra \W_\ra^\tkey+\b_\ra^\tkey, \\
\G_\id^\tquery&= \tbX^{(t)}_\id \W_\id^\tquery+\b_\id^\tquery, &\G_\ra^\tquery= \tbX^{(t)}_\ra \W_\ra^\tquery+\b_\ra^\tquery, \\
\G_\id^\tvalue&= \tbX^{(t)}_\id \W_\id^\tvalue+\b_\id^\tvalue, &\G_\ra^\tvalue= \tbX^{(t)}_\ra \W_\ra^\tvalue+\b_\ra^\tvalue,
\end{align}
where $\b$ is the bias term as used in the network layers, and we omit the subscript $^{(t)}$ under the clear context. Taking the projected features as inputs, we illustrate how to peel those unrelated components from the current branch. Below we take face identity distillation (the top block in Fig.~\ref{fig:ct_unit_framework}) as an example to introduce cross transformer.

To purify face identity representation, we expect to remove some identity-unrelated noises (here induced by race). Taking the current race components $\G_\ra^\tkey$ as the reference, we project the face representation $\G_\id^\tquery$ to the race space by computing the attention weight as follows
\begin{align}
	S_{i,j} = \frac{\exp( \frac{1}{\sqrt{d'}} [\G_\id^\tquery]_i [\G_\ra^\tkey]\tp_j )} {\sum^{d'}_{j=1} \exp(\frac{1}{\sqrt{d'}}[\G_\id^\tquery]_i [\G_\ra^\tkey]\tp_j)},
	\label{equ:race_bias_attention}
\end{align}
where $S_{i,j}$ measures the impact of $i$-{th} position in $\G_\id^\tquery$  on $j$-th position in $\G_\ra^\tkey$. Intuitively, $\G_\id^\tquery$ and $\G_\ra^\tkey$ are used to build the relationships between face and race features, thus the resultant $\S$ reveals the race-related/biased facial regions, which will be visualized in Section~\ref{sec:ablation_study} to investigate the race-related/-specific facial features qualitatively.
With the cross attention, spatial dependencies between any two positions of the face and race features can be captured. For the feature at a certain position, it is updated via aggregating features at all positions with weighted summation, where the attention weights are adaptively decided by the feature similarities between the corresponding two positions.

To further estimate the identity-unrelated component induced by race as formulated in Eqn.~(\ref{equ:epsilon_nonid}),  we transform the $S_{i,j}$ with an identity projection function as follows
\begin{align}
\bep_\ra^{(t)} =  \S\times G_\id^{\tvalue}(\tbX^{(t)}_\id),
\label{equ:race_bias}
\end{align}
where $G_\id^{\tvalue}$ denotes the identity projection, and $\times$ denotes the matrix multiplication. The output $\bep_{ra}^t$ corresponds to the identity-unrelated components induced by race, so we can obtain the less biased face representation at stage $t$ according to Eqn.~(\ref{equ:id_and_noise}), formally,
\begin{align}
	\X_{\id}^{(t)}= \tbX^{(t)}_{\id} - \bep_{\ra}^{(t)}.
\end{align}

To make cross transformer position-aware, we additionally resort to the relative distance encodings \cite{bello2019attention,  ramachandran2019stand, shaw2018self}. The position embedding takes into account the relative distances between features at different facial locations and effectively associates information across the face and race features with positional awareness.  Here we adopt the two dimensional relative position self-attention as used in the literature~\cite{shaw2018self}.

In the advanced self-attention method \cite{waswani2017attention, dosovitskiy2020image}, the input representation can be projected into different spaces via different learnable parameters, which fall into the  multi-head self-attention (MSA) paradigm. By projecting the input representation into several sub-spaces,  the model is capable of paying attention to different positions in the input representation. Inspired by this observation, we may extend the above cross transformer to the multi-head case. Let $H$ denotes the number of heads, the feature dimension $d'$ of each sub-space is set as: $d' = \frac{d}{H}$. The output is the linear transformation of concatenation of all the attention outputs:
\begin{align}
	\bep_{\ra}^{(t)} = \text{Concat}([\bep_{\ra}^{(t)}]_1, [\bep_{\ra}^{(t)}]_2, ..., [\bep_{\ra}^{(t)}]_H),
	\label{equ:race_bias_multi_head}
\end{align}
where $\text{Concat}$ denotes the concatenation operation in the feature dimension, $[\bep_{\ra}^{(t)}]_i$ means the output of the $i$-th head in MSA.

Conversely, we can remove some face identity information from the branch of noises/race, and obtain more identity-unrelated representation of noises (induced by race) in the bottom block as shown in Fig.~\ref{fig:ct_unit_framework}. With the cross attention mechanism in the proposed CT module, the identity-unrelated component induced by race could be washed out to some extent while the identity-related features would be reserved.
Unlike other fair face recognition approaches, PCT is deployed to merely distill the identity-related component from the identity-unrelated part induced by race. Thus, PCT is capable of suppressing the bias induced by race while enhancing the discriminability of face representations simultaneously.


\subsection{Progressive Learning}

In order to purify face representation in a top-down manner, the distilled face representation $\X_\id^{(t)}$ at $t$-th stage is fed into the input to the $t+1$ stage for the further refinement. To fuse the multi-layer convolutional network, we employ the standard convolutional operation to encode the input $\X_\id^{(t)}$ to high-level feature $\tbX_\id^{(t+1)}$, where the size of feature map could be reduced into one half when pooling is used. Formally, together with the noise representation, we have the formulas 
\begin{align}
	\tbX_{\id}^{(t+1)} = \text{conv}(\mathbf{X}_{\id}^{(t)}),~~
    \tbX_{\ra}^{(t+1)} = \text{conv}(\mathbf{X}_{\ra}^{(t)}),
\end{align}
where $\text{conv}$ means the standard convolutional operation within the $t$-th stage in the FR network, $t = \{1,2,\cdots, T-1\}$ denotes the index of the $T$ stages. With the evolution of face decoupling, more face identity information would be aggregated into the identity-related network (top branch in Fig.~\ref{fig:pct_framework}) while those identity-unrelated information of race factor would be suppressed progressively. At the last stage, we obtain the output decoupled features for face and race classification, as illustrated in the right part in Fig.~\ref{fig:pct_framework}.

\subsection{Loss Function}

Finally, we employ both the face recognition loss and race classification loss to learn the face and race representation in an end-to-end manner. For face recognition, we choose to use  ArcFace \cite{deng2019arcface} or CosFace \cite{wang2018cosface} for face recognition. The training objective of face identity can be formulated as:
\begin{align}
	\mathcal{L}_{\text{face}} &= -\frac{1}{N}\sum_{i=1}^{N}\log \frac{e^{(s(\cos(\theta_{y_i} + m)))}}{e^{(s(\cos(\theta_{y_i} + m)))}  + \sum_{j=1, j \neq y_i}^{n}e^{(s\cos\theta_j)}},\\
\theta_{y_i} &= \arccos\frac{\langle \w_{y_i}, \x_i\rangle}{\|\w_{y_i}\|\cdot\|\x_i\|},
	\label{equ:arcface_loss}
\end{align}
where $\x_i$ denotes the deep feature of the $i$-th training sample belonging to the $y_i$ classes, $\w_j$ is the $j$-th column of the weight in the last fully-connected classification layer, $N$ is the batch size, $\theta_{j}$ denotes the angle between weight $\w_j$ and feature $\x_i$, $s$ is a scale factor (64 as default), $m$ is an additive angular margin penalty between $\w_j$ and $\x_i$ to enhance the intra-class compactness and inter-class discrepancy.
We refer the interested reader to \cite{deng2019arcface} for more details and the explanation for the ArcFace loss. Besides, we may also use CosFace for face recognition whose details can be found in \cite{wang2018cosface}. The comparison of loss functions between ArcFace and CosFace is provided in the experiment part.
For race classification (RC),  we adopt the cross entropy loss $\mathcal{L}_{\text{race}}$. Finally, we integrate the two object functions for training,
\begin{align}
	\mathcal{L}_{\text{total}} = \mathcal{L}_{\text{face}}  + \alpha\mathcal{L}_{\text{race}},
	\label{equ:loss_total}
\end{align}
where the hyper-parameter $\alpha$ controls the importance of the RC term.
We present the thorough experimental analysis of the proposed PCT in the next section.

\section{Experiments}
In this section, we present the experimental evaluations of the proposed PCT.
Before showing the results, we will describe the experimental settings, including the training and evaluation datasets, the implementation details and the evaluation protocol.
Then, we compare our method with the state-of-the-art face recognition methods that aim to mitigate the bias. Finally, we provide the ablation analysis to verify the reasonability of the proposed PCT.

\subsection{Experimental setup}

\textbf{Datasets:} We employ the popular \textbf{BUPT-Balancedface} \cite{wang2020mitigating} and \textbf{BUPT-Globalface}  \cite{wang2020mitigating} datasets for training.
BUPTBalancedface contains 1.3M images of $28,000$ celebrities and is approximately race-balanced with $7,000$ identities per race.
BUPT-Globalface contains 2M images of $38,000$ celebrities, and its racial distribution is approximately the same as the real distribution of the world’s population.
The identities in the two datasets are grouped into 4 categories, i.e. Caucasian, Indian, Asian and African, according to their races.

For fairness testing, we employ the \textbf{RFW} \cite{wang2019racial} and \textbf{BFW} \cite{robinson2020face} datasets.
RFW dataset consists of four race subsets, namely Caucasian, Asian, Indian, and African. Each subset contains approximately 10K images of 3K individuals for face verification.
Compared with RFW, the BFW dataset contains balanced face images with more attributes, including identity, gender, and race. The identities in the BFW dataset are categorized into eight demographic groups according to two genders and four ethnic groups (i.e., Black, White, Asian, and India). Each demographic group in the BFW dataset consists of 200 subjects with $2,500$ face images.

\begin{table*}[htb]
	\centering
	\caption{Face verification performance (\%) of PCT, state-of-the-art fair face recognition methods on RFW (training dataset: BUPT-Balancedface). \textbf{Bold} denotes the best.}
	\label{tab:rfw_resnet34_balancedface1}
	\begin{tabular}{c|c|c|c|c|c|c|c}
		\hline
		& Methods (\%) & African  & Asian  & Caucasian  & Indian  & AVE ($\uparrow$) & STD ($\downarrow$) \\
		\hline
		\hline
		\multirow{2}{*}{\shortstack[l]{ResNet-34}} & DebFace \cite{gong2020jointly} & 93.67 & 94.33 & 95.95 & 94.78 &  94.68 & 0.83\\
		
		& PFE \cite{shi2019probabilistic} & 95.17 & 94.27 & \textbf{96.38} & 94.60 & 95.11 & 0.93 \\
		\hline
		\multirow{5}{*}{\shortstack[l]{ResNet-34, ArcFace}} &	Baseline \cite{deng2019arcface}  &  93.98 & 93.72 & 96.18 & 94.67 & 94.64 & 1.11  \\
		& MTL &  94.82  & 94.47  & \textbf{96.60}  & 95.23  & 95.28  & 0.93 \\
		& GAC \cite{gong2020mitigating} & 94.12  & 94.10  & 96.02  & 94.22  & 94.62  & 0.81 \\
		& RL-RBN \cite{wang2020mitigating} &  95.00  & 94.82  & 96.27  & 94.68  & 95.19  & 0.93 \\
		& \textbf{PCT (Ours)} & \textbf{95.72} & \textbf{94.98} & 96.22  & \textbf{95.33} & \textbf{95.56} & \textbf{0.53} \\

		\hline
		\multirow{4}{*}{\shortstack[l]{ResNet-34, CosFace}} & Baseline \cite{wang2018cosface} & 92.93 & 92.98 & 95.12 & 93.93 & 93.74 & 1.03 \\
		& MTL  & 95.20  & 94.58 & 96.82 & 95.60 & 95.55 & 0.94  \\
		& RL-RBN \cite{wang2020mitigating} & 95.27  & 94.52  & 95.47 & 95.15 & 95.10 & \textbf{0.41} \\
		
		& \textbf{PCT (Ours}) & \textbf{96.02} & \textbf{94.87} & \textbf{96.72}  & \textbf{96.02} & \textbf{95.91} & 0.77 \\
		\hline
		\hline
		\multirow{2}{*}{\shortstack[l]{ResNet-50, ArcFace}} & MTL & 96.05 & 95.25 & \textbf{97.20} & 96.05 & 96.14 & 0.80  \\
		& PCT (\textbf{Ours}) & \textbf{96.22} & \textbf{95.73} & 97.00  & \textbf{96.38} & \textbf{96.33} & \textbf{0.52} \\
		\hline
		\multirow{3}{*}{\shortstack[l]{ResNet-50, CosFace}} & MTL & 95.82 & 94.93 & 96.73 & 95.78 & 95.82  & 0.74   \\
		& GAC \cite{gong2020mitigating} & 94.77  & 94.87  & 96.20  & 94.98  & 95.21  & \textbf{0.58} \\
		
		& \textbf{PCT (Ours)} & \textbf{95.83} & \textbf{95.48} & \textbf{96.90}  & \textbf{96.12} & \textbf{96.08} & 0.60 \\
		\hline
	\end{tabular}
\end{table*}	

\begin{table*}[htb]
	\centering
	\caption{Face verification performance (\%) of PCT, state-of-the-art fair face recognition methods on RFW (training dataset: BUPT-Globalface). \textbf{Bold} denotes the best.}
	\label{tab:rfw_resnet34_globalface}
	\begin{tabular}{c|c|c|c|c|c|c|c}
		\hline
		& Methods (\%) & African & Asian & Caucasian & Indian & AVE ($\uparrow$) & STD ($\downarrow$)\\
		\hline
		\hline
		\multirow{4}{*}{\shortstack[l]{ResNet-34, ArcFace}} & Baseline \cite{deng2019arcface}  &  93.87 & 94.55 & 97.37 & 95.86 & 95.37 & 1.53  \\
		& MTL &  95.13  & \textbf{95.92}  & \textbf{97.92}  & 96.05  & 96.26  & 1.18 \\
		& RL-RBN \cite{wang2020mitigating}  &  94.87  & 95.57  & 97.08  & 95.63  & 95.79  & \textbf{0.93} \\
		& \textbf{PCT(Ours)} & \textbf{95.87} & 95.45 & 97.68  & \textbf{96.15} & \textbf{96.29} & 0.97 \\
		\hline
		\multirow{4}{*}{\shortstack[l]{ResNet-34, CosFace}} & Baseline \cite{wang2018cosface} & 92.17 & 93.50 & 96.63 & 94.68 & 94.25 & 1.90 \\
		& MTL & 95.07 & 95.53 & 97.87 & \textbf{96.52} & 96.25 & 1.24  \\
		& RL-RBN \cite{wang2020mitigating} & 94.27  & 94.58  & 96.03 & 95.15 & 95.01 & \textbf{0.77} \\
		& \textbf{PCT(Ours)} & \textbf{96.43} & \textbf{95.88} & \textbf{97.97}  & 96.50 & \textbf{96.70} & 0.89 \\
		\hline
		\hline
		\multirow{2}{*}{\shortstack[l]{ResNet-50, ArcFace}} & MTL & 96.30  & 	95.97 & 98.03 & 96.53 & 96.71 & 0.91   \\
		& \textbf{PCT(Ours)} & \textbf{96.58} & \textbf{96.05} & \textbf{98.15}  & \textbf{96.93} & \textbf{96.93} & \textbf{0.89} \\
		\hline
		\multirow{2}{*}{\shortstack[l]{ResNet-50, CosFace}}& MTL & 96.37 & 96.43  & 	\textbf{98.17}  & 96.95 & 96.98 & 0.83  \\
		& \textbf{PCT(Ours)} & \textbf{96.62} & \textbf{96.88} & 98.15  & \textbf{96.97} & \textbf{97.16} & \textbf{0.68} \\
		\hline
	\end{tabular}
\end{table*}

\textbf{Evaluation Protocol:}
For the RFW dataset, we follow the RFW face verification protocol with 6K pairs for each race. We utilize the average face verification accuracy (AVE) of four races as the metric to evaluate the total performance of the FR methods.
To measure the fairness of the proposed PCT and the compared FR methods, we exploit the standard deviation (STD) among different races.  STD reflects the amount of discrepancy among  various racial groups.

For the BFW dataset, we use FPR for the eight demographic groups (two genders, four races) to measure the overall performance of the fair FR methods. To measure the fairness, we exploit STD among the eight groups. We utilize the officially released face pairs and calculate the FPR for each group at a global cosine similarity threshold which is determined by the specified global FPR. 
We obtain the STD of the eight demographic groups on BFW dataset as:
$\delta = \frac{1}{K} \sqrt{\sum_{k=1}^{K}(\mathcal{F}_k - \mu)^2}$. $\mathcal{F}_k$ means the group-specific FPRs. $K$ means the number of groups, $\mu$ is the mean of the group-specific FPRs.
To normalize the scale for the group-specific FPRs, we divide $\delta$ by the specified global FPR.

\textbf{implementation details:} Following \cite{deng2019arcface, wang2020mitigating},  we cropped the face images to $112 \times 112$ with five facial landmarks detected by MTCNN \cite{zhang2016joint}. We normalize the face images by subtracting 127.5 and divided the pixel intensity by 128, then feeding the normalized face images to the face recognition and race classification networks.
For the race classification backbone, we adopt the ResNet18 network as that in \cite{gong2020mitigating}.
For the face recognition backbone, we exploit the ResNet-34 and ResNet-50 networks as they are popular network structures for face recognition. The ResNet-based networks consist of four stages, we embedded a CT module at each stage as shown in Fig.~\ref{fig:pct_framework}.

We adopt a batch-based stochastic gradient descent method to optimize the model. The base learning rates for FR and RC were set as 0.1 and 0.01. We scaled the learning rates by $0.1$ at $16, 24, 28$ epochs. The batch size was set as 256 for both the two classification tasks and the training process was finished at 32 epochs.
We set the momentum as 0.9 and the weight decay as 0.0005.
The optimal setting for the loss weight between the face and race classification tasks was set as $1:1$ by grid search.
The training of the models was completed on 8 NVIDIA Tesla V100 GPU with Pytorch framework \cite{paszke2017automatic}. For ResNet-34, it took about 10 hours to finish optimizing the PCT model on the BUPTBalancedface dataset 18 hours on the BUPT-Globalface dataset.
The number of heads $H$ in each CT module was set as 2 in our experiments. The selection of hyperparameter $H$ is discussed in Section \ref{sec:ablation_study}.
For the ArcFace and CosFace loss functions, we follow the common setting as that in \cite{deng2019arcface, wang2020mitigating} and set $s = 64$, then set $m = 0.35$ for ArcFace, $m = 0.35$ for CosFace.
The proposed PCT models were trained on BUPT-Balancedface or BUPT-Globalface datasets with ground truth race and identity labels.

\begin{figure*}[htb]
	\centering
	\includegraphics[width=0.8\linewidth]{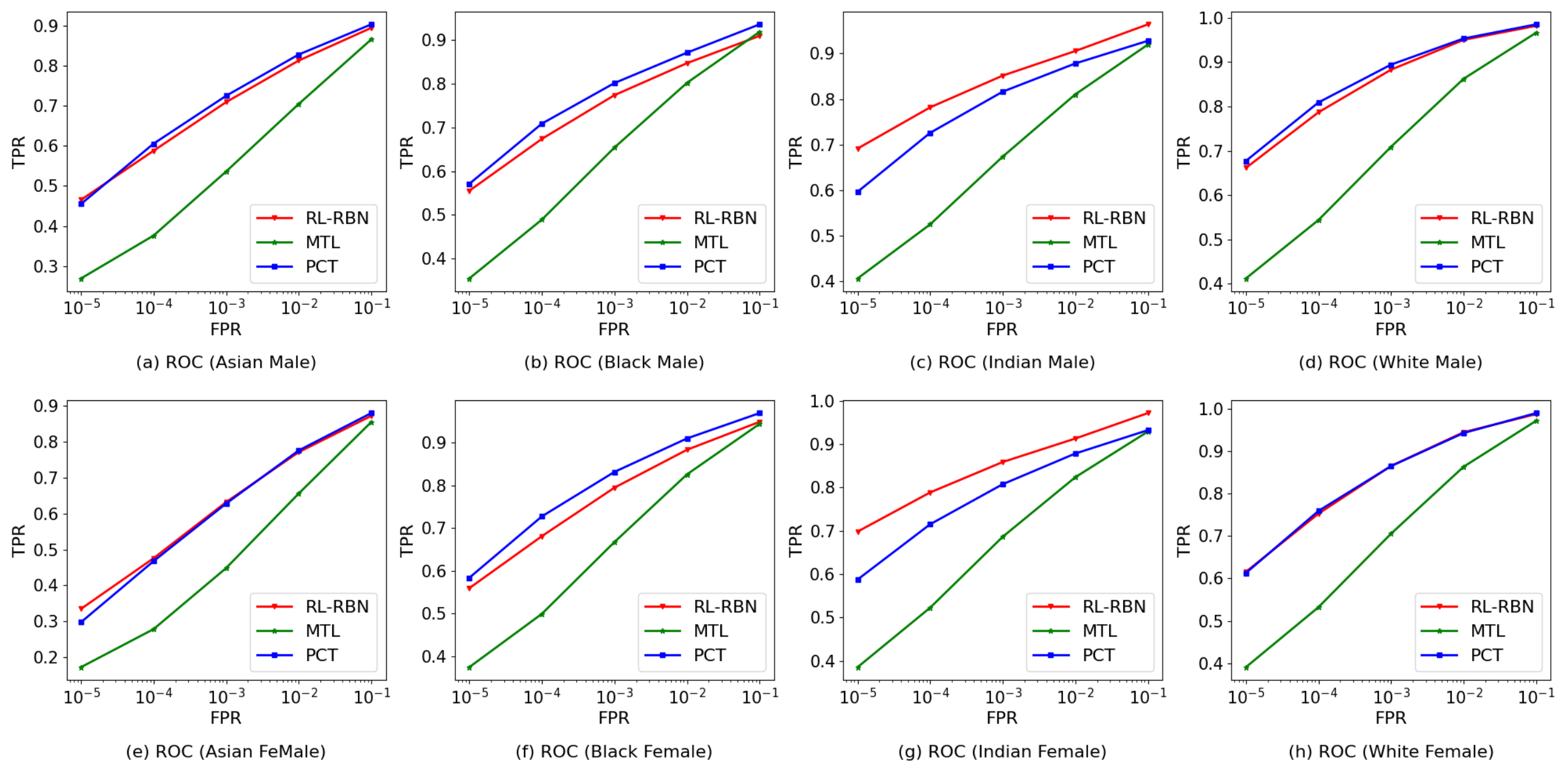}
	\caption{
		ROC curves for BFW (training dataset: BUPT-Balancedface). PCT outperforms or is comparable with other methods in the sub-figures (a), (b), (d), (e), (f), (h).
	}
	\label{fig:ROC_BFW_BalancedFace}
\end{figure*}

\begin{figure*}[htb]
	\centering
	\includegraphics[width=0.8\linewidth]{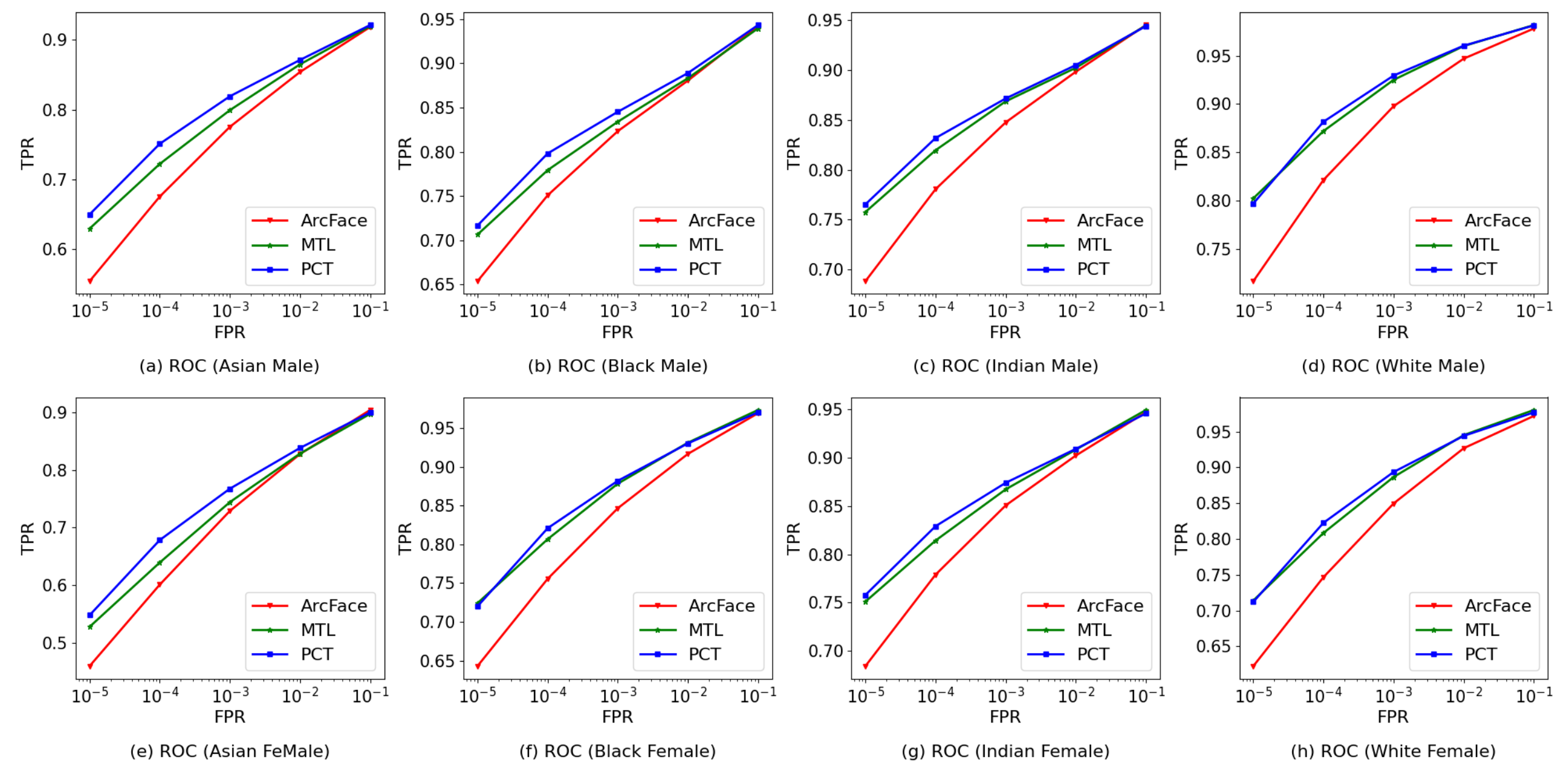}
	\caption{
		ROC curves for BFW (training dataset: BUPT-Globalface). PCT obtains slight improvements in all the sub-figures (a-h).
	}
	\label{fig:ROC_BFW_GlobalFace}
\end{figure*}

\begin{figure*}[htb]
	\centering
	\includegraphics[width=0.8\linewidth]{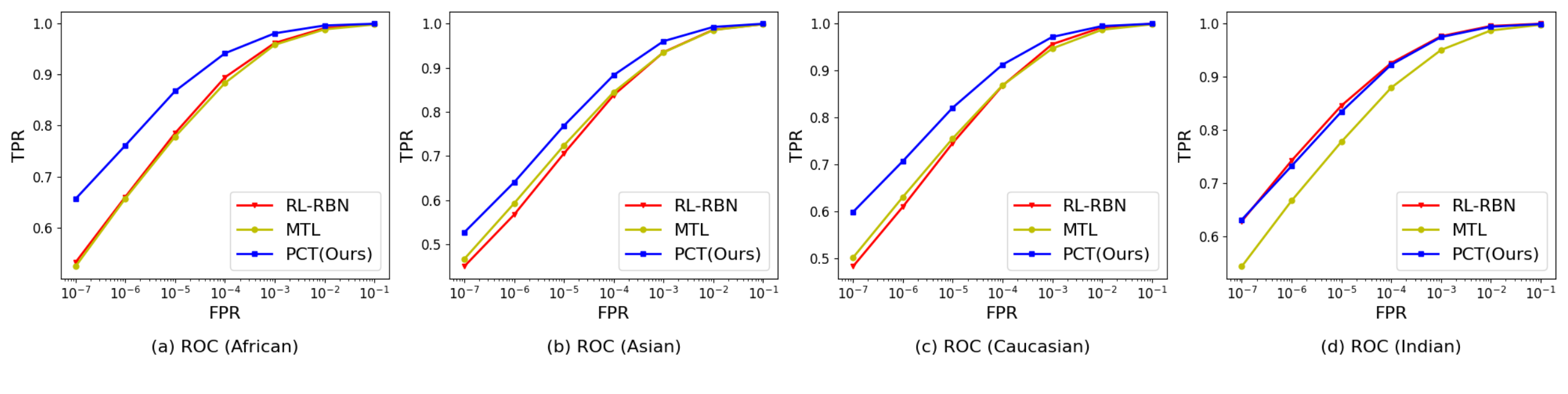}
	\caption{
		ROC curves for RFW dataset (training dataset: BUPT-Balancedface). PCT show its superiority in the sub-figures (a-c) and is comparable with the second best in the sub-figure (d).
	}
	\label{fig:ROC_RFW_BalanceFace}
\end{figure*}

\begin{figure*}[htb]
	\centering
	\includegraphics[width=0.8\linewidth]{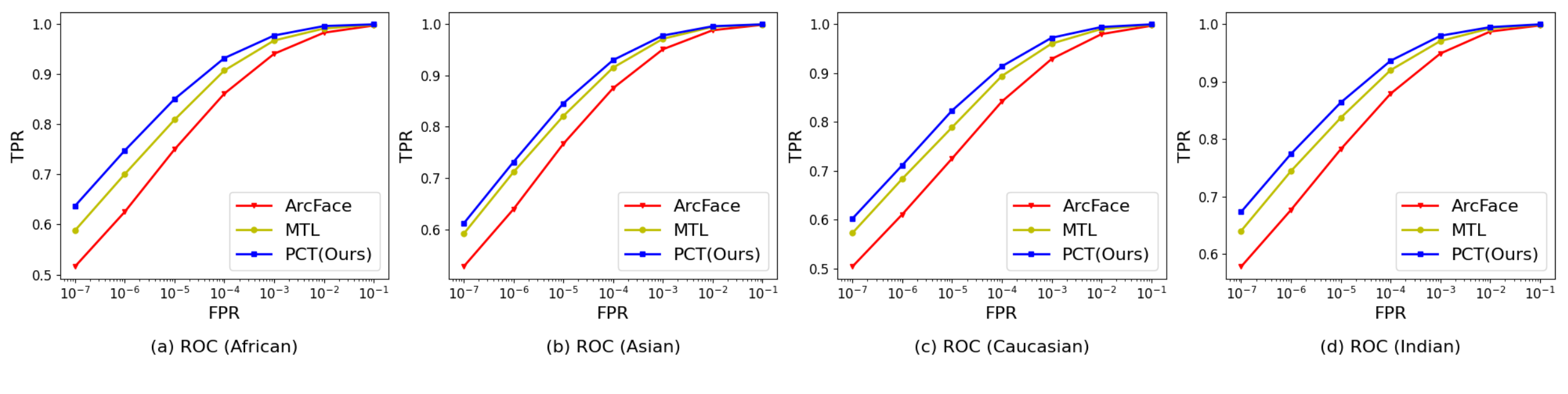}
	\caption{
		ROC curves for RFW (training dataset: BUPT-Globalface). PCT obtains slight improvements in all the sub-figures (a-d).
	}
	\label{fig:ROC_RFW_GlobalFace}
\end{figure*}

\subsection{Comparisons with state-of-the-art methods}
We compare PCT with other state-of-the-art fair face recognition methods, including the DebFace \cite{gong2020jointly}, Probabilistic Face Embeddings (PFE) \cite{shi2019probabilistic}, GAC \cite{gong2020mitigating}, RL-RBN \cite{wang2020mitigating}.
Among them, PFE \cite{shi2019probabilistic} represents each face image as a Gaussian distribution in the latent space. The mean of the distribution estimates the most likely feature values and the variance shows the uncertainty in the feature values. PFE is capable of handling the feature ambiguity dilemma for unconstrained face recognition.
Debface \cite{gong2020jointly} consists of a novel de-biasing adversarial network that learns to extract disentangled feature representations for both unbiased face recognition and demographics estimation.
GAC \cite{gong2020mitigating} mitigates the FR bias by using adaptive convolution kernels and attention mechanisms on faces based on their demographic attributes.
RL-RBN \cite{wang2020mitigating} contains a reinforcement learning-based race-balanced network that is trained to find the optimal margins for non-Caucasians.

We additionally compare PCT with a regular FR baseline which has the same architecture as the face recognition backbone in PCT using the ArcFace \cite{deng2019arcface} or CosFace \cite{wang2018cosface} loss function. Besides, we trained the face recognition and race classification tasks in a multi-task manner, where the two tasks share a classification backbone and have their respective classification heads. We term the multi-task learning method as MTL, as illustrated in Tab.~\ref{tab:rfw_resnet34_balancedface1}, \ref{tab:rfw_resnet34_globalface}, \ref{tab:bfw_resnet34_balanceface}, \ref{tab:bfw_resnet34_globalface}.

\subsubsection{Experimental results on RFW dataset}
Tab.~\ref{tab:rfw_resnet34_balancedface1} illustrates the face verification performance with the \textbf{BUPT-Balancedface} as the training and RFW as the evaluation dataset. The first column shows the loss function and network structure configurations of different FR approaches.

As illustrated Tab.~\ref{tab:rfw_resnet34_balancedface1}, our method performs better than the common baselines with the ResNet-based network structures, i.e., ArcFace \cite{deng2019arcface} , CosFace \cite{wang2018cosface}.
With ResNet-34, PCT outperforms ArcFace with 0.92\% improvements in the average face verification accuracy. When measuring the bias of the face representation, PCT significantly mitigates the racial bias and halves the standard deviation.
Although the methods \cite{deng2019arcface, wang2018cosface} are verified to obtain good performance for face recognition, they do not consider the demographic bias explicitly, thus the biases in data are inevitably transmitted to the FR models through network learning.
With ``ResNet-34 + ArcFace'' configuration, our proposed PCT obtains the best in both the bias and the average face verification performance.
Compared with DebFace \cite{gong2020jointly} that disentangles demographic attributes from the face representation, PCT shows higher face verification accuracies in all the four cohorts and lower standard deviation simultaneously. It is because race represents subject-specific intrinsic characteristics and is a  critical component that constitutes face patterns. Removing the race component from the face representation will reduce its discrimination \cite{gong2020jointly, liu2018exploring, lu2019experimental}.
Compared with GAC \cite{gong2020mitigating}, PFE \cite{shi2019probabilistic}, RL-RBN \cite{wang2020mitigating}, our proposed PCT achieves higher and fairer face verification performance and by mitigating the bias by cross transforming the face and race representation so as to suppress the identity-unrelated component induced by race gradually.
PCT also outperforms MTL because the MTL learns the face and race classification without extra constraint, thus the bias induced by race naturally remains in the learned face representation.
When it comes to ``ResNet-34 + CosFace'' configuration, the STD of our proposed PCT lags behind RL-RBN \cite{wang2020mitigating}. However, PCT outperforms RL-RBN in both the individual racial cohort and the average performance. It indicates the reduction of racial bias in our proposed PCT is obtained along with the accuracy improvements of all the four races.

We also trained the ResNet-50 models on BUPT-Balanceface dataset with our proposed PCT, as shown in the bottom part in Tab.~\ref{tab:rfw_resnet34_balancedface1}. It shows that both the individual accuracy and the average accuracy of PCT is  much better than that of GAC, while our standard deviation (0.60) is comparable with that of GAC (0.58). It reveals the proposed PCT is capable of equally representing the faces in different race cohorts while enhancing the discrimination of the learned face representation.
Compared with the conventional multi-task learning method (MTL), PCT shows consistent improvements in the face recognition accuracy and the equality of the learned face representation. It indicates the feasibility of the proposed CT modules in PCT that mitigates the bias induced by race gradually.

Tab.~\ref{tab:rfw_resnet34_globalface} illustrates the face verification performance
with the \textbf{BUPT-Globalface} as the training and RFW as the evaluation dataset.  Similar to  the aforementioned comparisons in Tab.~\ref{tab:rfw_resnet34_balancedface1}, we group the methods according to their loss functions and network structures in the first column.
Compared with other methods, our proposed PCT shows consistent high face verification accuracy and relatively low racial bias on various combinations of loss functions and network structures.
For example, with the ``ArcFace + ResNet-34" congfiguration,  PCT obtains 0.5\% improvments in the average accuracy and comparable standard deviation compared with RL-RBN \cite{wang2020mitigating}.
When it comes to ``CosFace + ResNet-34" configuration, the performance gap between PCT and RL-RBN \cite{wang2020mitigating} reaches 1.69\%.
The promotions of PCT verify the benefits of suppressing the identity-unrelated component induced by race in the learned face representation. Besides, the improvements of PCT over other compared methods are in line with the observation in \cite{wang2020mitigating} that colored faces are more susceptible to noise and image quality than Caucasians. Our proposed PCT mitigates the bias/noise induced by race with the proposed CT module in each stage in the classification network gradually, thus achieves balanced generalization ability of the learned face representation. The above results in Tab.~\ref{tab:rfw_resnet34_balancedface1}, \ref{tab:rfw_resnet34_globalface}  show that our method can achieve competitive performances on both race balanced and unbalanced datasets, with regard to the mean and standard deviation of accuracy.

From the experimental results in Tab.~\ref{tab:rfw_resnet34_balancedface1}, \ref{tab:rfw_resnet34_globalface}, we can conclude that our proposed PCT obtains the highest accuracy on Caucasian, and the lowest performance on Asian, respectively. This can be explained in two-fold: 1) Although the number of Asians is much larger than that of Indians and Africans in BUPT-Globalface dataset, this group still needs a larger margin because it is the most difficult race to recognize even with balanced training \cite{wang2020mitigating, terhorst2021comprehensive}. 2) There are many low-resolution face images in the Asian group in BUPT-Balanceface dataset, while this is not the case for the images in other races. In order to go deep into this phenomenon, we used the RetinaFace \cite{Deng_2020_CVPR} to detect the faces for the images in BUPT-Balancedface dataset, then calculated the average pixels in the facial regions according to the race annotations. For the African, Asian, Caucasian, India cohorts, the average pixels within the facial regions are 49335, 26395, 48119, 38487. It is clear that the face images in the Asian group suffer from low resolution compared with other racial groups, which can cause difficulty for accurate FR.

\begin{table}[htb]
	\centering
	\small
	\caption{Bias degree of protocol on BFW. Low bias degree mean fair face representation (training dataset: BUPT-Balancedface).}
	\label{tab:bfw_resnet34_balanceface}
	\begin{tabular}{c|c|c|c|c|c}
		\hline
		Methods (\%) & $10^{-5}$ & $10^{-4}$ & $10^{-3}$ & $10^{-2}$  & $10^{-1}$ \\
		\hline
		\hline
		
		ArcFace  & 0.86 & 0.72 & 0.58 & 0.42 & \textbf{0.21}   \\
		MTL & \textbf{0.49} & 0.64 & 0.54 & 0.47 & 0.33  \\
		\textbf{PCT (Ours)}  & 0.64 & \textbf{0.62} & \textbf{0.39} & \textbf{0.34} & 0.24   \\
		\hline
		CosFace & 0.86 & \textbf{0.44} & 0.36 & 0.33 & 0.25  \\
		MTL & 0.75 & 0.70 & 0.66 & 0.51 & 0.37 \\
		\textbf{PCT (Ours)}  & \textbf{0.58} & \textbf{0.44} & \textbf{0.34} & \textbf{0.30} & \textbf{0.22} \\
		\hline
	\end{tabular}
\end{table}

\begin{table}[htb]
	\centering
	\caption{Bias degree of protocol on BFW. Low bias degree mean fair face representation (training dataset: BUPT-Globalface).}
	\label{tab:bfw_resnet34_globalface}
	\begin{tabular}{c|c|c|c|c|c}
		\hline
		Methods (\%) & $10^{-5}$ & $10^{-4}$ & $10^{-3}$ & $10^{-2}$ & $10^{-1}$ \\
		\hline
		\hline
		ArcFace & 2.30 & 1.69 & 1.43 & 1.03 & 0.57  \\
		MTL  & \textbf{1.03} & 	\textbf{0.72} &	0.61 &	0.44 &	0.22   \\
		\textbf{PCT (Ours)} & \textbf{1.03}  &  0.76  &  \textbf{0.49}  & \textbf{ 0.34}  &  \textbf{0.15}  \\
		\hline
		CosFace & \textbf{0.63} &	0.74 &	0.58 &	0.41 &	0.21  \\
		MTL & 0.99 & 0.74 & \textbf{0.56} & 0.38 & 0.23    \\
		\textbf{PCT (Ours)} & 0.85  &  \textbf{0.71}  &  \textbf{0.56}  &  \textbf{0.40}  &  \textbf{0.19}  \\
		\hline
	\end{tabular}
\end{table}

\subsubsection{Experimental results on BFW}
The results are shown in Tab.~\ref{tab:bfw_resnet34_balanceface}, \ref{tab:bfw_resnet34_globalface}.
The experimental results show that our PCT shows consistent low bias than the compared methods at various FPRs. When we adopt the BUPT-Balanceface as the training dataset shown in Tab.~\ref{tab:bfw_resnet34_balanceface}, our PCT shows improved equality at FAR = 0.01, 0.001, and 0.0001.
When we set  BUPT-Globalface as the training data in Tab.~\ref{tab:bfw_resnet34_globalface}, our PCT shows low bias at FAR = 0.1, 0.01, 0.001.
These comparisons clearly and convincingly show that PCT can significantly improve the fairness of the learned face representation on both the race balanced and unbalanced datasets.

We additionally illustrate Receiver Operating Characteristic Curve (ROC) curves in Fig.~\ref{fig:ROC_BFW_BalancedFace}, \ref{fig:ROC_BFW_GlobalFace} on the eight groups in BFW dataset, and Fig.~\ref{fig:ROC_RFW_BalanceFace},  \ref{fig:ROC_RFW_GlobalFace} on the four racial groups in RFW  dataset with the ResNet-34 based FR models.
Compared with MTL and RL-RBN \cite{wang2020mitigating}, PCT obtains slight improvements in many racial groups, e.g., (a), (b), (d), (e), (f) in Fig.~\ref{fig:ROC_BFW_BalancedFace}. Compared with ArcFace and MTL, PCT shows its superiority in all the sub-figures (a-h) in Fig.~\ref{fig:ROC_BFW_GlobalFace}.
PCT also outperforms the compared methods in nearly all the sub-figures in Fig.~\ref{fig:ROC_RFW_BalanceFace}, \ref{fig:ROC_RFW_GlobalFace}.
MTL  outperforms ArcFace as race represents subject-specific intrinsic characteristics, explicitly learning face and race classification will enhance the discrimination of the facial features to some degree.
However, the race also induces bias into the facial representation.
To improve the robustness of features against race attribute, our proposed PCT suppresses the bias induced by race with the adaptive stage-wise CT modules and achieves both high FR accuracy and fairness simultaneously.

\begin{table}[htb]
	\centering
	\setlength{\tabcolsep}{4pt}
	\caption{Face verification performance on RFW dataset by adding none, one or stage-wise CT modules  (training dataset: BUPT-Balancedface). \textbf{Bold} denotes the best.}
	\label{tab:rfw_resnet34_head_balanceface}
	\begin{tabular}{c|c|c|c|c|c|c}
		\hline
		Methods (\%) & African & Asian & Caucasian & Indian & AVE ($\uparrow$) & STD ($\downarrow$) \\
		\hline
		\hline
		w/o CT   &  93.98 & 93.72 & 96.18 & 94.67 & 94.64 & 1.11  \\
		Stage 1  & 94.13 & 93.77 & 95.85 & 95.25 & 94.75 & 0.97  \\
		Stage 2 & 95.53 & 94.38 & 96.17 & \textbf{95.53} & 95.40 & 0.75 \\
		
		Stage 3  & 95.13 & 94.00 & 95.95 & 95.27 & 95.09 & 0.81 \\
		Stage 4  & 95.22  & 93.55  & 96.13  & 94.92  & 94.96  & 1.07 \\
		\textbf{PCT} & \textbf{95.72} & \textbf{94.98} & \textbf{96.22}  & 95.33 & \textbf{95.56} & \textbf{0.53} \\
		\hline
	\end{tabular}
\end{table}

\begin{table}[htb]
	\centering
	\setlength{\tabcolsep}{4pt}
	\caption{Face verification performance on RFW dataset by adding none, one or stage-wise CT modules (training dataset: BUPT-Globalface). \textbf{Bold} denotes the best.}
	\label{tab:rfw_resnet34_head_globalface}
	\begin{tabular}{c|c|c|c|c|c|c}
		\hline
		Methods (\%) & African & Asian & Caucasian & Indian & AVE ($\uparrow$) & STD ($\downarrow$) \\
		\hline
		\hline
		w/o CT & 93.87 & 94.55 & 97.37 & 95.86 & 95.37 & 1.53   \\
		stage 1 & 94.13 & 93.77 & 95.85 & 95.25 & 94.75 & \textbf{0.97}   \\
		Stage 2 & 95.57 & 95.38 & 97.82 & 96.05 & 96.21 & 1.11  \\	
		Stage 3  & 95.32 &	94.95 &	\textbf{97.83} &	96.05 &	96.04 &	1.28   \\
		Stage 4  & 95.23 & 94.53 &	97.68 &	95.97 &	95.85 &	1.35  \\
		\textbf{PCT} & \textbf{95.87} &	\textbf{95.45} &	97.68 &	\textbf{96.15} &	\textbf{96.29} &	 \textbf{0.97} \\
		\hline
	\end{tabular}
\end{table}

\begin{table}[htb]
	\centering
	\setlength{\tabcolsep}{4pt}
	\caption{Ablation study on RFW dataset. Verification performance (\%) of protocol on RFW with state-of-the-art methods. * means the values are reported in the original papers ([BUPT-Balancedface]).}
	\label{tab:rfw_resnet34_balanced_face_head}
	\begin{tabular}{c|c|c|c|c|c|c}
		\hline
		Methods (\%) & African & Asian & Caucasian & Indian & AVE ($\uparrow$) & STD ($\downarrow$) \\
		\hline
		\hline
		H = 1 & 95.23 & 93.92 & 96.45 & 94.42 & 95.13 & 1.04  \\
		\textbf{H = 2} & \textbf{95.72} & \textbf{94.98} & 96.22  & \textbf{95.33} & \textbf{95.56} & \textbf{0.53}  \\
		H = 4 & 94.67 &	93.40 &	\textbf{96.52} &	93.92 &	94.63 &	1.37  \\
		\hline
	\end{tabular}
\end{table}

\begin{table}[htb]
	\centering
	\setlength{\tabcolsep}{4pt}
	\caption{Ablation study on RFW dataset. Verification performance (\%) of protocol on RFW with state-of-the-art methods. * means the values are reported in the original papers ([BUPT-Globalface]).}
	\label{tab:rfw_resnet34_global_face_head}
	\begin{tabular}{c|c|c|c|c|c|c}
		\hline
		Methods (\%) & African & Asian & Caucasian & Indian & AVE ($\uparrow$) & STD ($\downarrow$) \\
		\hline
		\hline
		H = 1 & 94.83 &	94.55 &	97.28 &	95.33 &	95.50 &	1.23    \\
		\textbf{H = 2} & \textbf{95.87} &	\textbf{95.45} &	\textbf{97.68} &	\textbf{96.15} &	 \textbf{96.29} &	 \textbf{0.97}  \\
		H = 4 & 94.92 & 94.18 & 97.25 & 94.95 & 95.33 & 1.33   \\
		\hline
	\end{tabular}
\end{table}

\begin{figure*}[htb]
	\centering
	\includegraphics[width=1.0\linewidth]{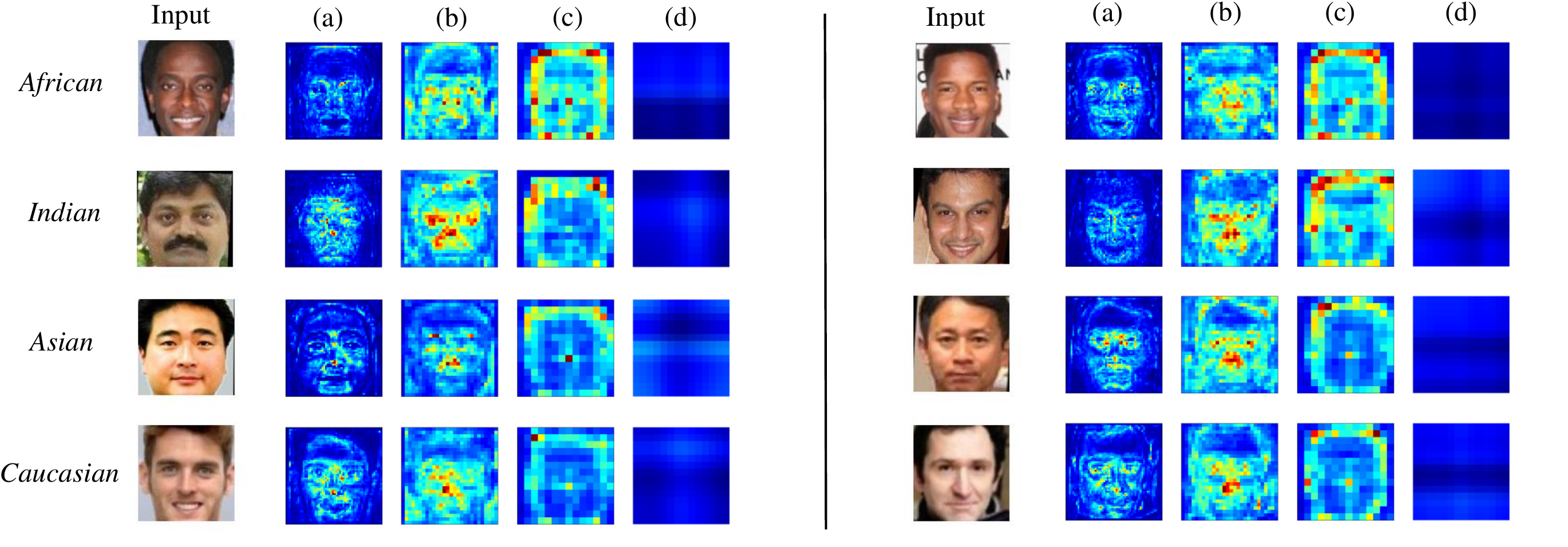}
	\caption{
		Attention maps of the four stage-wise cross transformer units. PCT perceives the race-related facial regions and suppresses the racial bias gradually. The sub-figures (a), (b), (c)  mainly covers the facial texture, facial features (e.g., eyes in African, Indian, nose in Asian, Caucasian), facial contour, respectively. This is in line with the observation in \cite{bulthoff2021predominance} that the three parts are the major determinants for different races. The deep red means the highly biased and race-related facial regions. The deep blue in (d) denotes the face representation contains less racial bias. Better viewed in color and zoom in.
	}
	\label{fig:race_per_subject}
\end{figure*}


\subsection{Ablation Study}
\label{sec:ablation_study}
To investigate the effectiveness of PCT, we conduct a quantitative evaluation to verify the contribution of the proposed CT module in each stage in our proposed PCT.
For a more detailed analysis, we also visualize the learned attention maps to reveal the biased facial regions related to race.

\textbf{Evaluation of stage-wise CT module:}
We analyze the performance variations of PCT by fully removing the race classification branch (w/o CT in Tab.\ref{tab:rfw_resnet34_head_balanceface}, \ref{tab:rfw_resnet34_head_globalface}) or only adding one CT module at certain stage in the classification network during the training process.
Tab.\ref{tab:rfw_resnet34_head_balanceface}, \ref{tab:rfw_resnet34_head_globalface} show the FR performance and bias variations.

As can be seen in Tab.\ref{tab:rfw_resnet34_head_balanceface}, \ref{tab:rfw_resnet34_head_globalface}, the baseline model without CT module shows the maximal standard deviation and it indicates the considerable bias in the learned face representation.
Adding one CT module at certain stage in the classification network is capable of decreasing the standard deviation and suppressing the bias induced by race.
It is worth noting that adding a CT module at the early stage (e.g., stage 1 or stage 1 in Tab.\ref{tab:rfw_resnet34_head_balanceface}, \ref{tab:rfw_resnet34_head_globalface},)  has more significant effect than adding it in the later stage (e.g., stage 3 or stage 4 in Tab.\ref{tab:rfw_resnet34_head_balanceface}, \ref{tab:rfw_resnet34_head_globalface}).
It might be explained faces in different racial groups usually have distinct facial texture distributions and morphological characteristics \cite{bulthoff2021predominance, fu2014learning}, and it has been verified that CNN tends to capture the low-level features such as color conjunctions, edges, corners in the early layers \cite{zeiler2014visualizing}.
Compared with other configurations, our proposed stage-wise progressive CT module (PCT) obtains higher and fairer face verification performance. The experimental results reveal the effectiveness of the stage-wise CT modules in the proposed PCT.

\textbf{Evaluation of different head numbers:} We analyze the performance variations of our method by exploiting different number of cross-attention heads $H$. As illustrated in Tab.~\ref{tab:rfw_resnet34_balanced_face_head},  \ref{tab:rfw_resnet34_global_face_head}, our proposed PCT always obtains good trade-off between FR accuracy and fairness with $H = 2$.
In Tab.~\ref{tab:rfw_resnet34_balanced_face_head}, PCT ($H = 2$) halves the standard deviation compared with PCT ($H = 1$).
In Tab.~\ref{tab:rfw_resnet34_balanced_face_head}, PCT ($H = 2$) nearly decreases one third standard deviation compared with PCT ($H = 1$). This is in line with the observation in \cite{waswani2017attention} that multi-head attention allows the model to jointly attend to information from different representation subspaces at different positions. As a comparison, the single attention head inhibits such benefits.
With $H = 4$, PCT shows decreased FR accuracy and increased bias. It might be explained that $H = 4$ incorporates too many trainable parameters. Applying  $H = 4$ to each CT module is prone to overfitting.
The comparisons in Tab.~\ref{tab:rfw_resnet34_balanced_face_head}, \ref{tab:rfw_resnet34_global_face_head} reveal $H = 2$ makes good compromise between the model capability and the amount of trainable parameters.

\textbf{Visualization:} To understand the learned fair face representation, we show the learned attention maps of some representative faces in Fig.~\ref{fig:race_per_subject}.
Although the individual differences exist in the learned attention maps, we can still observe some meaningful phenomenons from the consistencies within the same racial group and discrepancies between different races: (1) The columns (a), (b), (c) in Fig.~\ref{fig:race_per_subject}  mainly covers the facial texture, facial features (e.g., eyes in African, Indian, nose in Asian, Caucasian), and facial contour, respectively.
This phenomenon is in line with the observations in \cite{bulthoff2021predominance, fu2014learning} that facial texture, facial features (e.g., eyes, nose, mouth) and facial contour are the major determinants for different racial cohorts. It indicates the proposed PCT is capable of perceiving the race-related facial regions and suppresses the racial bias gradually.
(2)  When it comes to the fourth stage (column (d) in each row in Fig.~\ref{fig:race_per_subject}), the attention map tends to be overly similar with deep blue. It indicates the face representation in this stage contains less biased component induced by race.
It can be explained in two aspects. Firstly, the racial bias has been suppressed in the previous stages with the proposed CT modules.
Secondly, the racial bias mainly resides in the earlies stages in the FR model.
As evidenced in Tab.\ref{tab:rfw_resnet34_head_balanceface}, \ref{tab:rfw_resnet34_head_globalface}, the standard deviation of FR accuracies show its maximum when merely adding one CT module in the fourth stage among the stage-wise configurations.


\section{Conclusion}
Within this paper we have presented a transformer-based fair face recognition approach.
The proposed progressive cross transformer (PCT) suppresses the identity-unrelated component induced by race from the identity-related representation from a signal subspace decomposition perspective.
To distill the identity-related component and suppress the identity-unrelated part, we abstract the distillation process of the face representation as a signal denoising problem and propose a generalized parameterized model with the cross transformer (CT) mechanism.
This cross transformer mechanism can be applied in a stage-wise manner within a face recognition network so as to suppress the bias caused by race gradually.
Experimental results show our proposed PCT is capable of equally representing the faces in different race cohorts while enhancing the discrimination of the learned face representation.
For future work, we will explore how to suppress the bias caused by race or other demographic attributes in a self-supervised manner, as PCT relies on race annotation.

\ifCLASSOPTIONcaptionsoff
  \newpage
\fi

{
\bibliographystyle{IEEEtran}
\bibliography{ieee-abrv}
}

\begin{IEEEbiography}[{\includegraphics[width=1in,height=1.25in,clip,keepaspectratio]{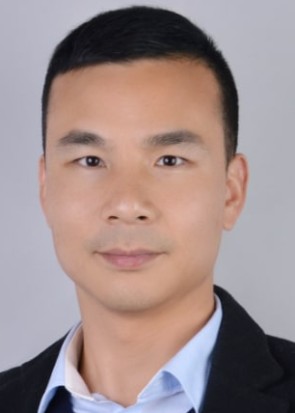}}]{Yong Li}  received the Ph.D. degree from Institute of Computing Technology (ICT), Chinese Academy of Sciences in 2020. He worked as a software engineer in Baidu company from 2015 to 2016. He has been a assistant professor at School of Computer Science and Engineering, Nanjing University of Science and Technology since 2020. His research interests include face-related deep learning, self-supervised learning and affective computing.
\end{IEEEbiography}

\begin{IEEEbiography}[{\includegraphics[width=1in,height=1.25in,clip,keepaspectratio]{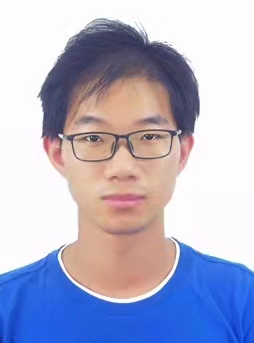}}]{Yufei Sun}  received the B.E. degree from Henan University, Kaifeng, China in 2019. He is working towards the M.S. degree in computer science and technology. His research interests include computer vision, fair face recognition.
\end{IEEEbiography}

\begin{IEEEbiography}[{\includegraphics[width=1in,height=1.25in,clip,keepaspectratio]{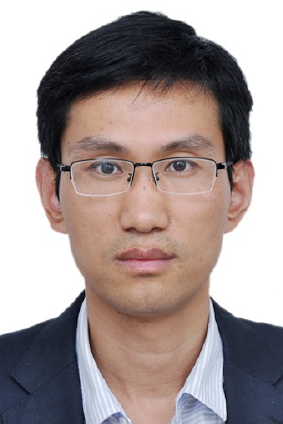}}]{Zhen Cui} received the Ph.D. degree from Institute of Computing Technology (ICT), Chinese Academy of Sciences in 2014. He was a Research Fellow in the Department of Electrical and Computer Engineering at National University of Singapore (NUS) from Sep 2014 to Nov 2015. He also spent half a year as a Research Assistant on Nanyang Technological University (NTU) from Jun 2012 to Dec 2012. Currently, he is a Professor of Nanjing University of Science and Technology, China.
	His research interests cover computer vision, pattern recognition and machine learning, especially focusing on  vision perception and computation, graph deep learning, etc.
\end{IEEEbiography}

\begin{IEEEbiography}[{\includegraphics[width=1in,height=1.25in,clip,keepaspectratio]{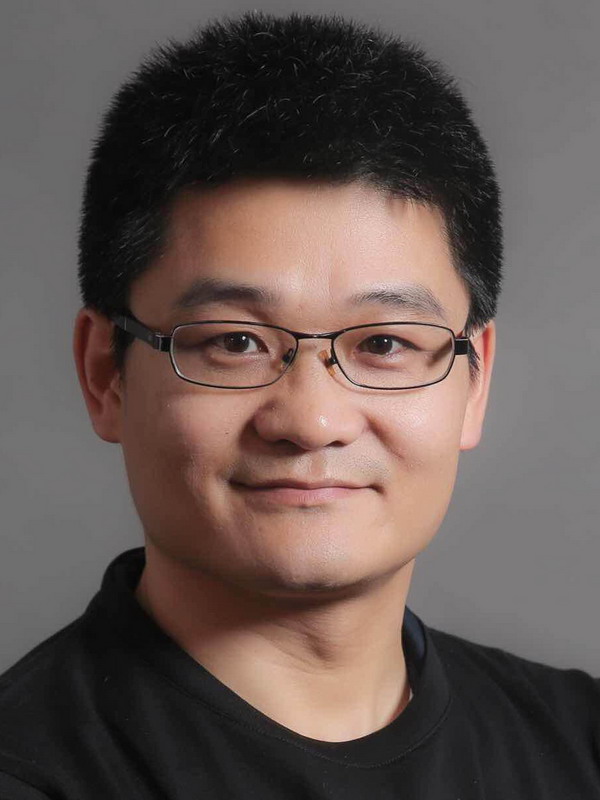}}]{Shiguang Shan} received M.S. degree in computer science from the Harbin Institute of Technology, Harbin, China, in 1999, and Ph.D. degree in computer science from the Institute of Computing Technology (ICT), Chinese Academy of Sciences (CAS), Beijing, China, in 2004. He joined ICT, CAS in 2002 and has been a Professor since 2010. He is now the deputy director of the Key Lab of Intelligent Information Processing of CAS. His research interests cover computer vision, pattern recognition, and machine learning. He especially focuses on face recognition related research topics. He has published more than 200 papers in refereed journals and proceedings in the areas of computer vision and pattern recognition. He has served as Area Chair for many international conferences including ICCV'11, ICPR'12, ACCV'12, FG'13, ICPR'14, ICASSP'14, ACCV'16, ACCV18, FG'18, and BTAS'18. He is Associate Editors of several international journals including IEEE Trans. on Image Processing, Computer Vision and Image Understanding, Neurocomputing, and Pattern Recognition Letters. He is a recipient of the China's State Natural Science Award in 2015, and the China’s State S\&T Progress Award in 2005 for his research work. He is also personally interested in brain science, cognitive neuroscience, as well as their interdisciplinary researche topics with AI.
\end{IEEEbiography}

\begin{IEEEbiography}[{\includegraphics[width=1in,height=1.25in,clip,keepaspectratio]{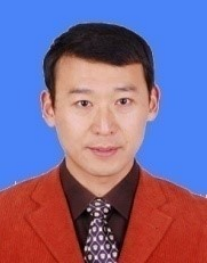}}]{Jian Yang}
	received the PhD degree from Nanjing University of Science and Technology (NUST), on the subject of pattern recognition and intelligence systems in 2002. In 2003, he was a postdoctoral researcher at the University of Zaragoza. From 2004 to 2006, he was a Postdoctoral Fellow at Biometrics Centre of Hong Kong Polytechnic University. From 2006 to 2007, he was a Postdoctoral Fellow at Department of Computer Science of New Jersey Institute of Technology. Now, he is a Chang-Jiang professor in the School of Computer Science and Technology of NUST. He is the author of more than 100 scientific papers in pattern recognition and computer vision. His journal papers have been cited more than 4000 times in the ISI Web of Science, and 9000 times in the Web of Scholar Google. His research interests include pattern recognition, computer vision and machine learning. Currently, he is/was an associate editor of Pattern Recognition Letters, IEEE Trans. Neural Networks and Learning Systems, and Neurocomputing. He is a Fellow of IAPR.
\end{IEEEbiography}

\end{document}